\documentclass[letterpaper]{article} 
\usepackage{aaai25}  
\usepackage{times}  
\usepackage{helvet}  
\usepackage{courier}  
\usepackage[hyphens]{url}  
\usepackage{graphicx} 
\usepackage{natbib}  
\usepackage{caption} 
\usepackage{algorithm}
\usepackage{algorithmic}
\usepackage{enumitem, lipsum, calc}
\usepackage{courier}
\usepackage{booktabs}
\usepackage{multirow}
\usepackage{titletoc}
\usepackage{amsmath}
\usepackage{amsfonts}
\usepackage{colortbl}
\usepackage[table]{xcolor}
\usepackage{cancel}

\newcommand{\proposed}{\textsf{RA-SGG}}

\usepackage{newfloat}
\usepackage{listings}
\DeclareCaptionStyle{ruled}{labelfont=normalfont,labelsep=colon,strut=off} 
\floatstyle{ruled}
\newfloat{listing}{tb}{lst}{}
\floatname{listing}{Listing}
\title{RA-SGG: Retrieval-Augmented Scene Graph Generation Framework via Multi-Prototype Learning}
\author{Kanghoon Yoon\textsuperscript{\rm 1}, Kibum Kim\textsuperscript{\rm 1}, Jaehyung Jeon\textsuperscript{\rm 1}, Yeonjun In\textsuperscript{\rm 1}, Donghyun Kim\textsuperscript{\rm 2}, Chanyoung Park\textsuperscript{\rm 1}\footnote{Corresponding author}}
\affiliations{
    \textsuperscript{\rm 1} Department of Industrial and Systems Engineering, Korea Advanced Institute of Science and Technology \\
    \textsuperscript{\rm 2} Department of Artificial Intelligence, Korea University \\

    \{ykhoon08, kb.kim, wogud405, yeonjun.in, cy.park\}@kaist.ac.kr, d\_kim@korea.ac.kr
}

\usepackage{bibentry}

\begin{document}

\maketitle

\begin{abstract}
Scene Graph Generation (SGG) research has suffered from two fundamental challenges: the long-tailed predicate distribution and semantic ambiguity between predicates. These challenges lead to a bias towards head predicates in SGG models, favoring dominant general predicates while overlooking fine-grained predicates. In this paper, we address the challenges of SGG by framing it as \emph{multi-label classification problem with partial annotation}, where relevant labels of fine-grained predicates are missing. Under the new frame, we propose \textbf{R}etrieval-\textbf{A}ugmented \textbf{S}cene \textbf{G}raph \textbf{G}eneration (\proposed), which identifies potential instances to be multi-labeled and enriches the single-label with multi-labels that are semantically similar to the original label by retrieving relevant samples from our established memory bank. Based on augmented relations (i.e., discovered multi-labels), we apply multi-prototype learning to train our SGG model. Several comprehensive experiments have demonstrated that~\proposed~outperforms state-of-the-art baselines by up to 3.6\% on VG and 5.9\% on GQA, particularly in terms of F@K, showing that~\proposed~effectively alleviates the issue of biased prediction caused by the long-tailed distribution and semantic ambiguity of predicates. The code of~\proposed~is available at \url{https://github.com/KanghoonYoon/torch-rasgg}.
\end{abstract}

%

\section{Introduction}

SGG is a pivotal task in scene understanding, aiming to detect objects and predict their relationships within an image. SGG models are adept at capturing rich visual information, and constructing scene graphs that encompass both object-level and relation-level information. This compositional representation of the scene graph is beneficial for various vision applications ~\cite{hildebrandt2020scene_graph_reasoning_vqa, schroeder2020scene_retrieval}. 
However, existing SGG approaches have encountered two fundamental challenges:

\smallskip
\noindent \textbf{1) Long-tailed Problem}: The predicate classes in benchmark SGG datasets, e.g., Visual Genome, follow a severely skewed distribution, where certain predicates are heavily represented (head classes) while others are sparsely annotated (tail classes). This challenge arises from annotators' tendency to label general predicates such as ``on'' more frequently than fine-grained predicates such as ``walking in.'' Consequently, the predictions of SGG models trained on datasets with such a skewed distribution tend to be biased towards head predicates, resulting in scene graphs dominated by general predicates. However, these scene graphs lack nuanced descriptions provided by tail predicates, which offer detailed relationship information.
\textbf{2) Semantic Ambiguity}: Another challenge stems from the ambiguous semantic boundaries between predicate classes. For a relation instance, predicates such as ``on,'' ``walking on,'' and ``walking in'' may be hard-to-distinguish as these predicates share semantic meanings, and thus SGG models need to capture nuanced visual cues to distinguish between them. However, learning the detailed visual differences is challenging as the tail predicates, such as ``walking in,'' rarely appear in the dataset whose predicate class distribution is usually skewed. Addressing these challenges is crucial for generating informative scene graphs with fine-grained predicates.

\begin{figure*}[t]
    \centering
    \includegraphics[width=1.6\columnwidth]{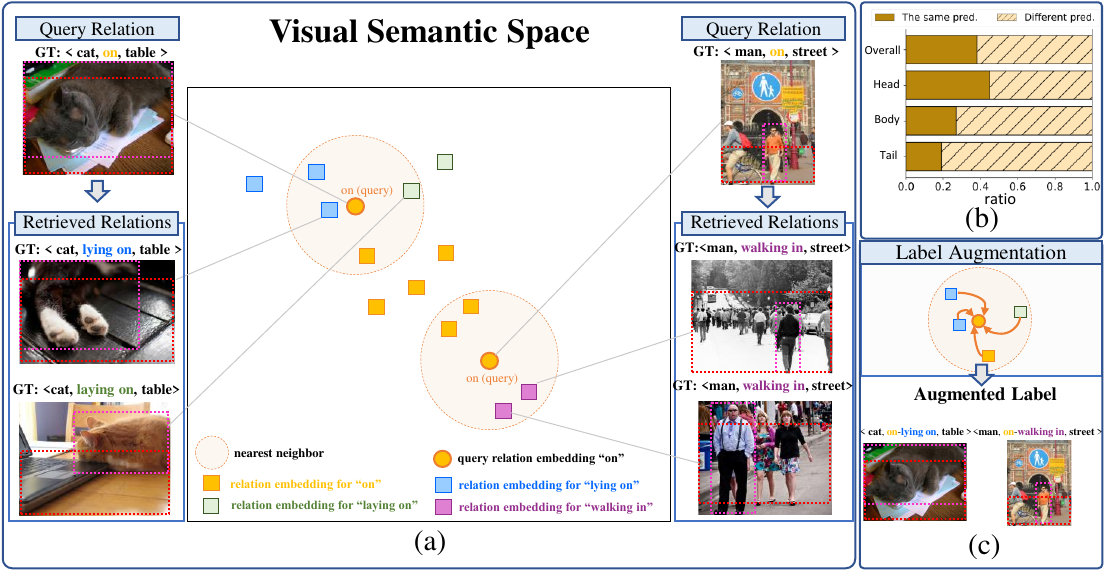}
    \caption{Illustration of the label augmentation of~\proposed~in SGG. (a) Examples of the query relation instances and their retrieved relation instances in the embedding space. \textbf{``GT''} denotes the ground truth. (b) The ratio of retrieved instances that have the same/different predicate as the query relation in the training data.
    (c) Fine-grained predicate label augmentation.}
    \label{fig:intro}
    \vspace{-2.3ex}
\end{figure*}

Seminal works have focused on generating fine-grained scene graphs by alleviating the issues posed by these challenges. Re-sampling, re-weighting, and various debiasing methods~\cite{Li2021bgnn,Lyu_2022_fgpl, kaihua2020tde, Kim_2024_CVPR} are utilized to generate the fine-grained scene graphs. Most recently,~\cite{li2022nice, zhang2022ietrans, Yu_2023_ICCVcacao} propose the data enhancement approaches by re-annotating the general predicates to fine-grained ones. However, this refinement process results in a trade-off, where comprehension of general predicates is diminished. 
Moreover, we find that fundamental challenges persist in the prior framework. Current SGG models rely on single-label classification, where predicates compete against each other during prediction, forcing the model to select just one predicate while suppressing others. This approach is problematic because it ignores the nuanced nature of natural language, where multiple predicates convey the same single relationship. Hence, we cast the single-label classification with \emph{multi-label classification with partial annotation problem}~\cite{Ben-Baruch_2022cvpr_multilabel_partial_annot}. That is, we discover potential fine-grained predicates in the original training data and augment semantically similar predicates for relationships.


In this work, we propose a \textbf{R}etrieval-\textbf{A}ugmented \textbf{S}cene \textbf{G}raph \textbf{G}eneration (\proposed) framework that enhances single-label annotations to multi-labels with fine-grained predicates, addressing the challenges of the long-tailed predicate distribution and the semantic ambiguity. The core idea is to retrieve potential predicates that share semantics with the original predicate and use them as pseudo-labels for the relation, as shown in Figure~\ref{fig:intro}. 
Our approach is motivated by observations in the visual semantic space. Using the pre-trained SGG model, PE-Net~\cite{Zheng_2023_CVPR_PENET}, for nearest neighbor retrieval on training instances, we discovered that query relations and their retrieved instances often share semantic meanings despite having different predicate labels. For instance, a relation with the ground-truth predicate ``on'' retrieves different fine-grained predicates: $\{$``lying on'', ``laying on''$\}$  for $\langle$cat,on,table$\rangle$ or $\{$``walking in'', ``walking in''$\}$ for $\langle$man, on, street$\rangle$. Notably, this pattern persists across head, body, and tail predicate classes (Figure~\ref{fig:intro}.b), suggesting that \emph{a relation instance that is currently annotated with a single predicate might actually co-reference multiple latent predicates with similar meanings.}


Building on these observations,~\proposed~aims to discover latent predicates of a relation instance by retrieving semantically similar relation instances from a memory bank. We introduce three key mechanisms: (1) \emph{Reliable Multi-Labeled Instance Selection}: naively assigning pseudo-labels for all the relation instances would aggravate reliability. To obtain reliable pseudo-labels, we introduce a label inconsistency score which measures the discrepancy between the ground-truth predicate and its retrieved relation instances. (2) \emph{Unbiased Augmentation of Multi-Labels}: we introduce the inverse propensity score-based sampling strategy to sample the less frequent (i.e., fine-grained) predicates among the retrieved relation instances. (3) \emph{Multi-Prototype Learning with Multiple Predicates}: it ensures relation instance embeddings capture both original and discovered fine-grained predicate semantics. Our comprehensive experiments demonstrate that~\proposed~outperforms state-of-the-art SGG models, particularly in terms of F@K, which is a metric that emphasizes the ability to achieve high performance in fine-grained classes (i.e., tail classes) while minimally sacrificing the performance of general classes (i.e., head classes). \textbf{Our contribution} can be summarized as follows: 

\begin{enumerate}[leftmargin=*, topsep=1mm]
\item We highlight the problem of single-label classification in SGG and reformulate this in the light of \emph{multi-label classification with partial annotation problem}. 
\item We propose a novel retrieval-augmented SGG framework (\proposed), which simultaneously alleviates the long-tailed distribution and semantic ambiguity issues by discovering the latent predicates of a relation instance annotated with a single predicate. 
\item Our extensive studies demonstrate that~\proposed~achieves the state-of-art performance in terms of F@K, implying that~\proposed~effectively generates fine-grained scene graphs without sacrificing the understanding of general predicates.
\end{enumerate}

\section{Related Works}

\subsection{Scene Graph Generation}

SGG has garnered attention due to its capacity for dense understanding of a scene, bridging two modalities: vision and language. 
Early works have focused on capturing the contextual information within a scene using message-passing network and transformer architecture~\cite{zellers2018neuralmotif,graph_rcnn,Li2021bgnn,Shit2022Relformer}. 
Recent SGG studies have been centered on improving performance for fine-grained predicates (i.e., tail classes) to obtain informative scene graphs. However, it is challenging to predict fine-grained predicates due to the fact that the predictions of SGG models easily collapse into general and trivial predicates, suffering from both the long-tailed distribution and semantic ambiguity issues. To address this,~\cite{Li2021bgnn,Lyu_2022_fgpl} focus training on tail predicate classes via resampling/reweighting, and \cite{kaihua2020tde,dong2022stacked,dpl_jeon} proposed debiasing methods to mitigate the bias towards head classes. However, merely amplifying the weight of tail classes can lead to overfitting of SGG models to the limited number of tail instances~\cite{zhai2022understanding_rwt}, as these methods do not contribute to increasing variations in the data for tail classes~\cite{2022improving_sgg_contrastve}.
As an alternative, approaches that enhance the training data have emerged. Specifically, NICE~\cite{li2022nice} identifies noisy predicates in data by using a pre-trained model, and IE-Trans~\cite{zhang2022ietrans} replaces general predicates with fine-grained ones or uses fine-grained predicates to fill in missing annotations in the training data. ST-SGG~\cite{kim2024adaptive} assigns pseudo-labels to the missing annotated instances.

However, the data enhancement approaches eliminate general predicates from the dataset, leading to a significant performance drop on these general predicates. This strategy overlooks a comprehensive understanding of general predicates to instead obtain fine-grained scene graph. Moreover, we argue that adjusting data based on a single prediction of model is risky in the SGG nature, where the models are trained on the extremely long-tailed distribution. To resolve these issues, we addresses the SGG problem in the light of the multi-label classification problem for comprehensive understanding across all predicate classes. Moreover,~\proposed~does not rely on a single model prediction but considers the predicate distribution of the retrieved relation instances of a target relation to accurately assign predicates as a multi-label.


\subsection{Retrieval Augmented Generation}

Retrieval Augmented Generation (RAG) is a widespread approach in natural language processing (NLP). RAG employs a non-parametric memory to supplement intrinsic knowledge of a language model by retrieving relevant passages from knowledge bases~\cite{pmlr-v162-borgeaud22a_retro, guu2020icml_realm, lewis2020nips_rag_knowledge}. Similarly, in the realm of image synthesis, approaches like~\cite{casanova2021instanceconditioned,NEURIPS2022_ra_diffusion} refer to examples of near neighbors to elevate the quality of generated images. 
Furthermore, RAG verifies the ability to handle rare examples within the training dataset. Re-Imagen~\cite{chen2023reimagen}~effectively generates the rare entities by referencing images with similar text descriptions.~\cite{Long_2022_CVPR_longtail_retrieve}~enhances the image classification performance on tail classes by drawing upon image features from a memory bank.
In this work, we propose the retrieval-augmented scene graph generation framework to discover the latent multi-labels in the training dataset. Our retrieval process is performed on the established memory bank populated with the relation embedding, which is produced by a pre-trained SGG model. Our work is the first work, which shows the usefulness of retrieval-based augmentation for the SGG task.

\section{Preliminaries}
\label{section:Preliminaries}
\subsection{Scene Graph Generation Task}
We start by presenting the scene graph generation problem, which includes the detection of objects and the prediction on the classes of objects and their relationships. Formally, given an image, we aim at generating a scene graph $\mathbf{G}=\langle \mathbf{s}_i, \mathbf{p}_i, \mathbf{o}_i \rangle_{i=1}^N$, where $N$ is the number of triplets in the image. $\mathbf{s}_i$ represents the subject, which includes the bounding box position and the object class, while $\mathbf{o}_i$ represents the object, which is similarly defined. $\mathbf{p}_i$ represents the predicate between $\mathbf{s}_i$ and $\mathbf{o}_i$, and includes the predicate class.

\subsection{Visual Semantic Embedding Space in SGG}
PE-Net~\cite{Zheng_2023_CVPR_PENET}~proposes a novel SGG learning strategy, which reduces the distance between the prototype and visual relation embedding in the visual semantic space. This paradigm encourages an SGG model to discern between hard-to-distinguish predicates by measuring the distance between the relation embedding and its predicate class prototype. 

Formally, let $\boldsymbol{v_s}\in \mathbb{R}^{d}$ and $\boldsymbol{v_o}\in\mathbb{R}^{d}$ denote the visual semantic features of subject and object, respectively, which are extracted from both the visual feature and the word embedding of entity classes. We use $\boldsymbol{u_p}\in \mathbb{R}^{d}$ to denote the relation embedding extracted from the visual features of the union box, and $\boldsymbol{t_p}\in \mathbb{R}^{{d'}}$ to denote the word embedding of the predicate class. The core idea of prototype learning is to match the subject and object instances to the predicate instances by
$\mathcal{F}(\boldsymbol{v_s}, \boldsymbol{v_o}) \approx \mathbf{W}_p \boldsymbol{t_p} + \boldsymbol{u_p}$, where $\mathcal{F}$ is a fusion layer~\cite{Zheng_2023_CVPR_PENET} and $\mathbf{W}_p \in \mathbb{R}^{d \times d'}$ is a feature transformation matrix. To implement this, PE-Net considers the language semantic of predicates as a prototype (i.e., $\mathbf{W}_p \boldsymbol{{t}_p}$), and makes the relation embedding (i.e., $\boldsymbol{{r}} = \mathcal{F}(\boldsymbol{v_s}, \boldsymbol{v_o}) - \boldsymbol{u_p}$) close to its corresponding the predicate prototype. The prototype learning loss is defined as:
{\small
\begin{align}
\label{eqn:penet}
    \mathcal{L}_{\text{proto}} &= - \log \frac{\exp(\langle \boldsymbol{\bar{r}}, \boldsymbol{\bar{c}_{\text{gt}}} \rangle) / \gamma}{ \sum_{j=1}^{N_p} \exp(\langle \boldsymbol{\bar{r}}, \boldsymbol{\bar{c}}_j \rangle) / \gamma}, \,\,\,  \\ 
    & \text{where} \quad \boldsymbol{\bar{r}}=\textsf{Proj}(\boldsymbol{{r}})
    \text{and}  \,\, \,\, \boldsymbol{\bar{c}}_j=\textsf{Proj}(\mathbf{W}_p \boldsymbol{{t}_p}) \nonumber
\end{align}
}
\noindent where $N_p$ is the number of predicate categories, $\gamma$ is a learnable temperature hyperparameter and \textsf{Proj} is a 2-layer MLP shared across both the relation embedding $\boldsymbol{r}$ and the prototypes. $\boldsymbol{\bar{c}_{\text{gt}}}$ denotes the class prototype of the ground truth predicates.
This prototype-guided learning has demonstrated the effectiveness of producing discriminative relation embeddings in the visual semantic space, compared to the conventional classifier-based SGG models. 
Hence, our proposed method,~\proposed, utilizes the relation embeddings of the training instances learned from the above objective to instantiate the retrieval-augmented approache for SGG.

\begin{figure*}[t]
    \centering
    \includegraphics[width=1.6\columnwidth]{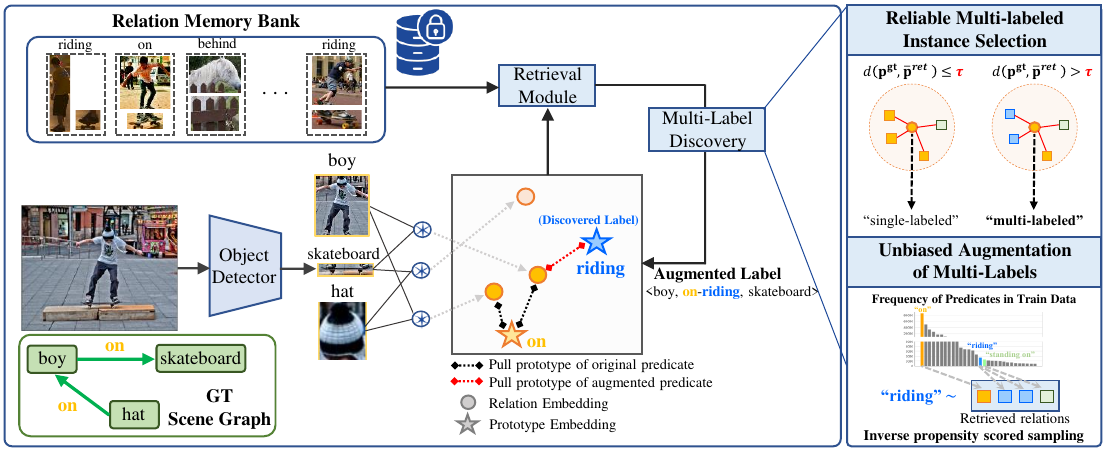}
    \vspace{-1ex}
    \caption{
    \proposed~first uses the relation embeddings for querying instances in the memory bank to retrieve visually and semantically similar instances. Then, it augments multi-labels by assigning pseudo-labels to potentially multi-labeled instances.
    }
    \label{fig:architecture}
    \vspace{-3.ex}
\end{figure*}

\section{Retrieval Augmented SGG}

In this section, we introduce~\proposed, a novel method that utilizes a retrieval process to discover and augment the latent and fine-grained predicates. We begin by formulating the SGG task as the multi-label classification with a partial (i.e., single) annotation problem, which aims to model the true unbiased predicate distribution based on the partially annotated samples. 
Subsequently, we explain how~\proposed~identifies relation instances that include potential multi-labels and augments the labels of these instances by assigning semantically similar predicates as additional pseudo-labels. Lastly, we outline the training procedure of~\proposed, which employs a multi-prototype learning strategy to effectively learn from the augmented relation labels. The pipeline of~\proposed~is illustrated in Figure~\ref{fig:architecture}.


\subsection{Multi-Label Classification with Partial Annotation}
\label{subsection:formulation}

Recall that existing SGG studies have relied on a single annotation for each relationship, overlooking the potential predicates that share similar semantic meanings. This oversight limits the model's ability to represent complex visual scenes, particularly affecting the representation of less common or tail predicates due to insufficient data variation. Herein, we first formulate the SGG task with multi-labeled classification problem with partial annotation to exploit the latent fine-grained predicates. 


Let $\mathbf{y}^*\in \{0, 1\}^{N_p}$ denote the ground-truth label vector of a relation instance from an unbiased true distribution, which is an ideal distribution without any latent unobserved annotations. $\mathbf{y}^*$ accommodates multi-labels including semantically relevant predicates, i.e., $\sum_{i=1}^{N_p}\mathbf{y}_i \geq 1$. In a common SGG setting, we deal with partially observed annotation with the observed label vector $\mathbf{y} \in \{0, 1\}^{N_p}$, where $\mathbf{y}$ is single-labeled or not labeled, i.e., $\sum_{i=1}^{N_p}\mathbf{y}_i = 1$ or $\sum_{i=1}^{N_p}\mathbf{y}_i = 0$. Our goal is to minimize the loss function $\mathcal{L}(\mathbf{y^*}, \mathbf{\hat{y}})$ computed on the samples from the unbiased true distribution by utilizing observed predicate labels $\mathbf{y}$, where $\mathbf{\hat{y}}$ is the probability predicted by an SGG model. 
However, since obtaining samples from the unbiased true distribution is not feasible, we approximate the loss using the available observed samples. Hence, we introduce the following inverse propensity-scored loss:
{\small
\begin{equation}
    \mathcal{L}_{ips} = - \sum_{i=1}^{N_p} \underbrace{  {{P(\mathbf{y}_i=1|\mathbf{y}_i^*=1}})^{-1}}_{\text{inverse propensity score}} \mathbf{y}_i \log \mathbf{\hat{y}}_i
    \label{eqn:ips_loss}
\end{equation}
}
\noindent where $\mathcal{L}_{ips}$ is the weighted cross entropy with the inverse propensity for each predicate class, and the propensity score is the label frequency~\cite{jain2016kdd_extreme_multi_label,chiou2021mm_dlfe}, which is the fraction of observed predicates among entire predicates in the unbiased true distribution. For simplicity, we follow \cite{jain2016kdd_extreme_multi_label} and use the frequency of each predicate class in the training dataset as the propensity score. 
Note that optimizing $\mathcal{L}_{ips}$ is approximately minimizing the loss using samples from the unbiased true distribution since $\mathcal{L}_{ips}$ serves as an unbiased estimator of the loss function $\mathcal{L}(\mathbf{y^*}, \mathbf{\hat{y}})$. However, directly minimizing the inverse propensity scored loss for SGG models is still restricted as it does not increase the variation of data instances of tail predicate classes, which incurs overfitting to a small number of tail predicates. Instead of optimizing the above $\mathcal{L}_{ips}$, we choose to augment a relation instance by sampling the predicate labels based on the inverse propensity score~\cite{cikm22_tailmixup}, and assign the sampled predicate labels to the relation instance as pseudo-labels. We will discuss this in detail in Section~\ref{subsection:retag}.

\subsection{Retrieval-Augmented Scene Graph Generation}
\label{subsection:retag}

We adopt a retrieval process to effectively train an SGG model based on the partially observed predicates. Specifically, we discover and augment the latent and fine-grained predicates of a relation instance by referring to the predicates of the retrieved instances. The process begins by retrieving semantically similar samples from the memory bank. We then leverage the predicates associated with these retrieved samples to identify relation instances that could be labeled with multiple predicates and to augment our dataset with these multi-labels. Although our method can be integrated into any SGG model, this paper specifically showcases its application through PE-Net as the backbone.

\smallskip

\noindent \textbf{Relation Retrieval Module.} \@ 
The relation retrieval module finds semantically similar relation instances based on the relation embedding. Specifically, a frozen memory bank is established using a pre-trained SGG model prior to the training of~\proposed. We store key-value pairs $(k_1, v_1), ..., (k_M, v_M)$ in the memory bank, where the key is the relation embedding ($\boldsymbol{{r}}$ in Eqn.~\ref{eqn:penet} for PE-Net), and the value is an associated one-hot vector representing the predicate class. Our approach limits memory storage upto $10$ entries for each unique triplet, which only utilizes 8\% of the training data. Note that the memory bank construction is executed during the pre-processing phase, without imposing additional computational costs in the training phase of~\proposed.

During the training phase,~\proposed~produces the relation embedding $\boldsymbol{{r}}$ for each subject-object pair in the training batch. The relation embeddings act as queries to retrieve $K$ neighbor instances from the memory bank based on the cosine similarity. Among the retrieved instances, we exclude the first instance to avoid the trivial case of retrieving itself.

\smallskip

\noindent \textbf{Reliable Multi-Labeled Instance Selection.} \@ Since some relation instances contain only a single semantic meaning, naively assigning pseudo-labels to all relation instances would aggravate reliability. Given the $K$ nearest neighbor instances $\{(\boldsymbol{r}_1^{\text{ret}}, \mathbf{p}_1^{\text{ret}}), ..., (\boldsymbol{r}_K^{\text{ret}}, \mathbf{p}_K^{\text{ret}}) \}$ of the query relation,~\proposed~first identifies whether the query relation is a single-labeled instance with only one predicate, or a multi-labeled instance that, despite potentially qualifying for multiple labels, currently has only one predicate. This selection is based on the label inconsistency between the ground-truth predicate of query relation $\textbf{p}^{\text{q}}$ and the averaged one-hot predicates of retrieved instances $\bar{\textbf{p}}^{\text{ret}}=\frac{1}{K} \sum_{i=1}^K \mathbf{p}_i^{\text{ret}}$. By computing the label inconsistency distance $d(\cdot, \cdot)$ for all query relation instances in the training batch, we obtain the $\mathcal{D}_{\text{single}}$ and $\mathcal{D}_{\text{multi}}$ as follows:
{\small
\begin{align}
\label{eqn:identification}
    & \mathcal{D}_{\text{single}} \leftarrow \{(\mathbf{s}_i, \mathbf{p}_i, \mathbf{o}_i)|d(\textbf{p}^{\text{q}}, \bar{\textbf{p}}^{\text{ret}})<\tau,\forall(\mathbf{s}_i, \mathbf{p}_i, \mathbf{o}_i)\in\mathcal{D}_{\text{Tr}} \} \\ 
    & \mathcal{D}_{\text{multi-}} \leftarrow \{(\mathbf{s}_i, \mathbf{p}_i, \mathbf{o}_i)|d(\textbf{p}^{\text{q}}, \bar{\textbf{p}}^{\text{ret}}) \geq \tau,  \forall(\mathbf{s}_i, \mathbf{p}_i, \mathbf{o}_i)\in\mathcal{D}_{\text{Tr}} \} \nonumber
\end{align}
}
where $\tau$ is the threshold hyperparameter and $d$ is the distance metric. If the given ground-truth predicate of the query relation is significantly different from that of the nearest neighbor samples, i.e., large inconsistency, then the query relation is likely to co-reference additional potential predicate labels among the predicates of the neighbor samples.

\smallskip

\noindent \textbf{Unbiased Augmentation of Multi-Labels.} 
Having identified potential relation instances to be multi-labeled, we augment the fine-grained predicates by assigning specific multi-labels to the training instances within $\mathcal{D}_{\text{multi}}$. In this augmentation stage, we aim to obtain the augmented labels that follow the unbiased true distribution by adopting the inverse propensity score-based sampling. Specifically, we obtain the sampling weight $w\in\mathbb{R}^{N_p}$ by aggregating the inverse propensities of the retrieved predicates, i.e., $w= \textsf{Softmax}(\sum_{k=1}^K s_k^{\text{ret}} \textbf{p}_k^{\text{ret}})$, where ${s}_1^{\text{ret}}$, ..., ${s}_K^{\text{ret}} \in \mathbb{R}$ are the inverse propensities of retrieved predicates, which are pre-computed as the inverse of the predicate frequency in the training dataset. For each triplet $(\mathbf{s}_i, \mathbf{p}_i, \mathbf{o}_i)$ in $\mathcal{D}_{\text{multi}}$, we augment the predicate labels using the sampling weight as:
{\small
\begin{align}
    \mathbf{p}_i \leftarrow \lambda_i \mathbf{p}_i^{\text{gt}} + (1-\lambda_i) \mathbf{p}_i^{\text{aug}}, \text{where} \:\: \mathbf{p}_i^{\text{aug}} \sim \textsf{Multinomial}(w)
\end{align}
}
\noindent where $\lambda_i\sim\textsf{Beta}(\alpha, \beta)$ controls the degree of the mixing coefficient. Our augmentation strategy generates pseudo-labels, considering not only the overall semantic meaning of nearest neighbors but also the frequency of predicates. This approach effectively leverages partial annotations to create multi-labeled instances, which can address the long-tailed distribution problem and semantic ambiguity in SGG.

\begin{table*}[t]
\centering
\scalebox{.61}{
\begin{tabular}{c|l|ccc|ccc|ccc}
\hline
\multirow{2}{*}{\textbf{B}} & \multirow{2}{*}{\textbf{Methods}} & \multicolumn{3}{c}{\textbf{Predicate Classification}} & \multicolumn{3}{c}{\textbf{Scene Graph Classification}} & \multicolumn{3}{c}{\textbf{Scene Graph Detection}} \\
 & & \cellcolor[HTML]{EFEFEF}R@50/100 & mR@50/100 & F@50/100 & \cellcolor[HTML]{EFEFEF}R@50/100 & mR@50/100 & F@50/100 & \cellcolor[HTML]{EFEFEF}R@50/100 & mR@50/100 & F@50/100  \\
 \hline
& KERN\cite{chen2019knowledge}{\tiny{CVPR'19}} & \cellcolor[HTML]{EFEFEF}65.8/67.6 & 17.7/19.2 & 27.9/29.9 & \cellcolor[HTML]{EFEFEF}36.7/37.4 & \; 9.4/10.0 & 15.0/15.8 & \cellcolor[HTML]{EFEFEF}27.1/29.8 & 6.4/7.3 & 10.4/11.7 \\
& BGNN\cite{Li2021bgnn}{\tiny{CVPR'21}} &  \cellcolor[HTML]{EFEFEF}59.2/61.3 & 30.4/32.9 & 40.2/42.8 & \cellcolor[HTML]{EFEFEF}37.4/38.5 & 14.3/16.5 & 20.7/23.1 & \cellcolor[HTML]{EFEFEF}31.0/35.8 & 10.7/12.6 & 15.9/18.6 \\
& DT2ACBS\cite{Desai_2021_ICCV}{\tiny{ICCV'21}} & \cellcolor[HTML]{EFEFEF}23.3/25.6 & 35.9/\textbf{39.7} & 28.3/31.1 & \cellcolor[HTML]{EFEFEF}16.2/17.6 & \textbf{24.8/27.5} & 19.6/21.5 & \cellcolor[HTML]{EFEFEF}15.0/16.3 & \textbf{22.0/24.0} & 17.8/19.4 \\
& HL-Net\cite{Lin_2022_CVPR_hlnet}{\tiny{CVPR'22}} &  \cellcolor[HTML]{EFEFEF}67.0/68.9 & \:\:\: - /22.8 & \:\:\: - /34.3 & \cellcolor[HTML]{EFEFEF}42.6/43.5 & \:\:\: - /13.5 & \:\:\: - /20.6 & \cellcolor[HTML]{EFEFEF}33.7/38.1 & \:\:- /9.2 & \:\: - /14.8 \\
& HetSGG\cite{yoon2023hetsgg}{\tiny{AAAI'23}} &   \cellcolor[HTML]{EFEFEF}57.8/59.1 & 31.6/33.5 & 40.9/42.8 & \cellcolor[HTML]{EFEFEF}37.6/38.7 & 17.2/18.7 &  23.6/25.2 & \cellcolor[HTML]{EFEFEF}30.0/34.6 & 12.2/14.4 & 17.3/20.3 \\
\multicolumn{1}{l|}{\multirow{-5.5}{*}{\rotatebox{90}{Specific}}} & SQUAT\cite{Jung_2023_CVPRsquat}{\tiny{CVPR'23}} &  \cellcolor[HTML]{EFEFEF}55.7/57.9 & 30.9/33.4 & 39.7/42.4 & \cellcolor[HTML]{EFEFEF}33.1/34.4 & 17.5/18.8 & 22.9/24.3 &\cellcolor[HTML]{EFEFEF} 24.5/28.9 & 14.1/16.5 & 17.9/21.0 \\
 \hline
& Motif\cite{zellers2018neuralmotif}{\tiny{CVPR'18}}  & \cellcolor[HTML]{EFEFEF}64.6/66.0 & 15.2/16.2 & 24.6/26.0 & \cellcolor[HTML]{EFEFEF}38.0/38.9 & 8.7/9.3 & 14.2/15.0 & \cellcolor[HTML]{EFEFEF}31.0/35.1 & 6.7/7.7 & 11.0/12.6  \\
& TDE\cite{kaihua2020tde}{\tiny{CVPR'20}} &  \cellcolor[HTML]{EFEFEF}46.2/51.4 & 25.5/29.1 & 32.9/37.2 & \cellcolor[HTML]{EFEFEF}27.7/29.9 & 13.1/14.9 & 17.8/19.9 & \cellcolor[HTML]{EFEFEF}16.9/20.3 & 8.2/9.8 & 11.0/13.2 \\
& DLFE\cite{chiou2021mm_dlfe}{\tiny{MM'21}} &  \cellcolor[HTML]{EFEFEF}52.5/54.2 & 26.9/28.8 & 35.6/37.6 & \cellcolor[HTML]{EFEFEF}32.3/33.1 & 15.2/15.9 & 20.7/21.5 & \cellcolor[HTML]{EFEFEF}25.4/29.4 & 11.7/13.8 & 16.0/18.8 \\
& NICE\cite{li2022nice}{\tiny{CVPR'22}} &  \cellcolor[HTML]{EFEFEF}55.1/57.2 & 29.9/32.3 & 38.8/41.3 & \cellcolor[HTML]{EFEFEF}33.1/34.0 & 16.6/17.9 & 22.1/23.5 & \cellcolor[HTML]{EFEFEF}27.8/31.8 & 12.2/14.4 & 17.0/19.8 \\
& GCL\cite{dong2022stacked}{\tiny{CVPR'22}} &  \cellcolor[HTML]{EFEFEF}42.7/44.4 & 36.1/38.2 & 39.1/41.1 & \cellcolor[HTML]{EFEFEF}26.1/27.1 & 20.8/21.8 & 23.2/24.1 & \cellcolor[HTML]{EFEFEF}18.4/22.0 & 16.8/19.3 & 17.6/20.6 \\
& IETrans\cite{zhang2022ietrans}{\tiny{ECCV'22}} & \cellcolor[HTML]{EFEFEF}54.7/56.7 & 30.9/33.6 & 39.5/42.2 & \cellcolor[HTML]{EFEFEF}32.5/33.4 & 16.8/17.9 & 22.2/23.3 & \cellcolor[HTML]{EFEFEF}26.4/30.6 & 12.4/14.9 & 16.9/20.0 \\
& CFA~\cite{Li_2023_ICCV_cfa}{\tiny{ICCV'23}} &  \cellcolor[HTML]{EFEFEF}54.1/56.6 & 35.7/38.2 &  43.0/45.6 & \cellcolor[HTML]{EFEFEF}34.9/36.1 & 17.0/18.4 & 22.9/24.4 & \cellcolor[HTML]{EFEFEF}27.4/31.8 & 13.2/15.5 & 17.8/20.8 \\
\multicolumn{1}{l|}{\multirow{-8.5}{*}{\rotatebox{90}{Motif}}} & ST-SGG\cite{kim2024adaptive}{\tiny{ICLR'24}} &  \cellcolor[HTML]{EFEFEF}53.9/57.7 & 28.1/31.5 &  36.9/40.8 & \cellcolor[HTML]{EFEFEF}33.4/34.9 & 16.9/18.0 & 22.4/23.8 & \cellcolor[HTML]{EFEFEF}26.7/30.7 & 11.6/14.2 & 16.2/19.4 \\ \hline

& $\text{PE-Net}^{\dagger}$\cite{Zheng_2023_CVPR_PENET}{\tiny{CVPR'23}} & \cellcolor[HTML]{EFEFEF}64.9/67.2 & 31.5/33.8 &  42.4/45.0 & \cellcolor[HTML]{EFEFEF}39.4/40.7 & 17.8/18.9 & 24.5/25.8 & \cellcolor[HTML]{EFEFEF}30.7/35.2 & 12.4/14.5 & 17.7/20.4 \\
& $\text{IETrans}^{\dagger}$\cite{zhang2022ietrans}{\tiny{ECCV'22}} & \cellcolor[HTML]{EFEFEF}49.3/51.8 & 33.5/36.0 &  39.9/42.5 & \cellcolor[HTML]{EFEFEF}31.2/32.3 & 18.3/19.4 & 23.1/24.2 & \cellcolor[HTML]{EFEFEF}24.2/28.4 & 13.7/16.2 & 17.5/20.6 \\
& $\text{CFA}^{\dagger}$\cite{Li_2023_ICCV_cfa}{\tiny{ICCV'23}} &  \cellcolor[HTML]{EFEFEF}57.8/61.6 & 30.0/33.2 &  39.5/43.1 & \cellcolor[HTML]{EFEFEF}36.2/37.1 & 15.9/18.2 & 22.1/24.4 & \cellcolor[HTML]{EFEFEF}25.6/29.8 & 14.4/17.1 & 18.4/21.7 \\ \cmidrule{2-11}

\multicolumn{1}{l|}{\multirow{-4.}{*}{\rotatebox{90}{PE-Net}}} & \proposed  & \cellcolor[HTML]{EFEFEF}62.2/64.1 & \textbf{36.2}/39.1 & \textbf{45.7}/\textbf{48.6} & \cellcolor[HTML]{EFEFEF}38.2/39.1 & 20.9/22.5 & \textbf{27.0}/\textbf{28.6} & \cellcolor[HTML]{EFEFEF}26.0/30.3 & 14.4/17.1 & \textbf{18.5}/\textbf{21.9}
\end{tabular}
}
\vspace{-2ex}
\caption{Performance (\%) of state-of-the-art SGG models on Visual Genome~\cite{krishna2017visualgenome}. F@K is the harmonic mean of mR@50/100 and R@50/100. $\dagger$ denotes the result produced by us using their official code.}
\label{tab:main_vg}
\end{table*}

\begin{table*}[t!]
\centering
\vspace{-1ex}
\scalebox{.61}{
\begin{tabular}{l|ccc|ccc|ccc}
\hline
\textbf{Models} &  \multicolumn{3}{c}{\textbf{Predicate Classification}} & \multicolumn{3}{c}{\textbf{Scene Graph Classification}} & \multicolumn{3}{c}{\textbf{Scene Graph Detection}} \\
 & \cellcolor[HTML]{EFEFEF}R@50/100 & mR@50/100 & F@50/100 & \cellcolor[HTML]{EFEFEF}R@50/100 & mR@50/100 & F@50/100 & \cellcolor[HTML]{EFEFEF}R@50/100 & mR@50/100 & F@50/100  \\
 \hline
VTransE\cite{Hudson_2017_CVPR_vtranse}{\tiny{CVPR'17}}        & \cellcolor[HTML]{EFEFEF}55.7/57.9 & 14.0/15.0 & 22.4/23.8 & \cellcolor[HTML]{EFEFEF}33.4/34.2 & 8.1/8.7   & 13.0/13.9 & \cellcolor[HTML]{EFEFEF}27.2/30.7 & 5.8/6.6   & 9.6/10.9  \\
Motif\cite{zellers2018neuralmotif}{\tiny{CVPR'18}}          & \cellcolor[HTML]{EFEFEF}65.3/66.8 & 16.4/17.1 & 26.2/27.2 & \cellcolor[HTML]{EFEFEF}34.2/34.9 & 8.2/8.6   & 13.2/13.8 & \cellcolor[HTML]{EFEFEF}28.9/33.1 & 6.4/7.7   & 10.5/12.5 \\
VCTree\cite{tang2019vctree}{\tiny{ICCV'19}}          & \cellcolor[HTML]{EFEFEF}63.8/65.7 & 16.6/17.4 & 26.4/27.5 & \cellcolor[HTML]{EFEFEF}34.1/34.8 & 7.9/8.3   & 12.8/13.4 & \cellcolor[HTML]{EFEFEF}28.3/31.9 & 6.5/7.4   & 10.6/12.0 \\
$\text{PE-Net}^{\dagger}$ \cite{Zheng_2023_CVPR_PENET}{\tiny{CVPR'23}}           & \cellcolor[HTML]{EFEFEF}54.3/56.0   & 26.2/27.1 & 35.4/36.5 & \cellcolor[HTML]{EFEFEF}26.2/27.0 & 11.2/11.5 & 15.7/16.1 & \cellcolor[HTML]{EFEFEF}19.5/22.9 & 10.3/11.9 & 13.5/15.7 \\
\hline
PE-Net+\proposed    & \cellcolor[HTML]{EFEFEF}48.3/50.1 & \textbf{35.4/36.8} & \textbf{40.9}/\textbf{42.4} & \cellcolor[HTML]{EFEFEF}19.9/20.8 & \textbf{16.4/17.2} & \textbf{18.0}/\textbf{18.8} & \cellcolor[HTML]{EFEFEF}16.3/19.0 & \textbf{12.9/15.0} & \textbf{14.4}/\textbf{16.8}
\end{tabular}
}
\vspace{-2ex}
\caption{Performance (\%) of state-of-the-art SGG models on GQA~\cite{Hudson_2019_CVPR_gqa}}
\label{tab:main_gqa}
\vspace{-1ex}
\end{table*}

\begin{figure*}[t!]
    \begin{minipage}{0.85\columnwidth}
    \centering
    \includegraphics[width=0.99\columnwidth]{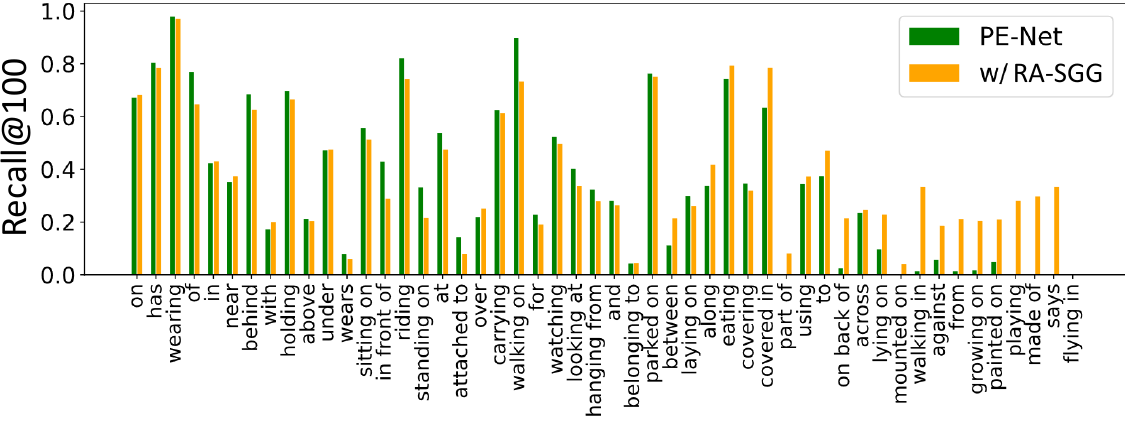}
    \caption{Per predicate comparison of~\proposed~with PE-Net on VG. The task is PredCls.}
    \label{fig:perclass_vg}
    \end{minipage}
    \hfill
    \begin{minipage}{1.2\columnwidth}
    \centering
    \includegraphics[width=0.987\columnwidth]{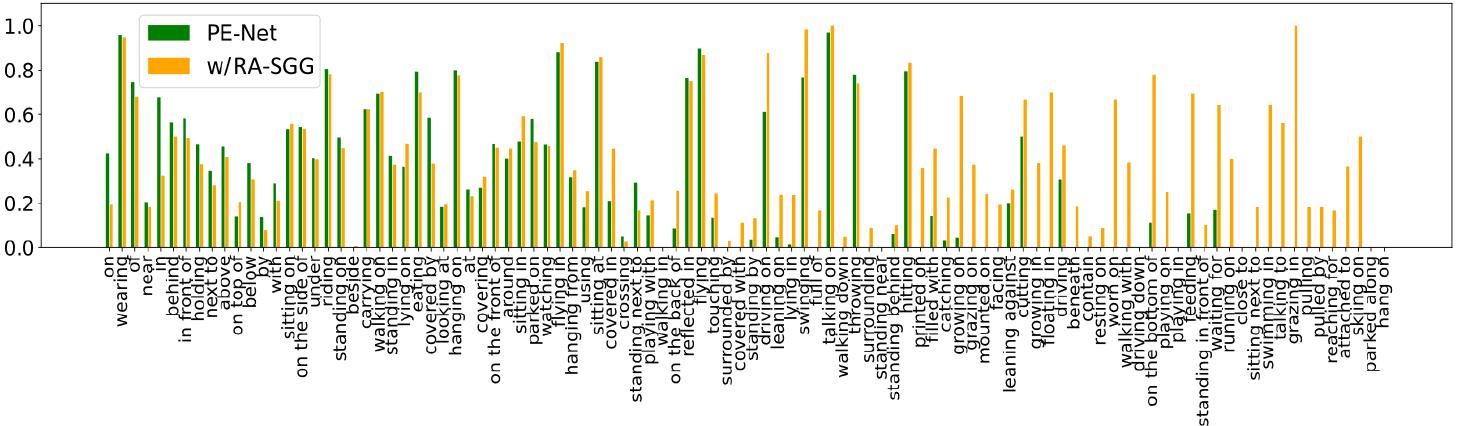}
    \vspace{-1.3ex}
    \caption{Per predicate comparison of~\proposed~with PE-Net on GQA. The task is PredCls. The predicates are sorted by the frequency. 
    }
    \label{fig:perclass_gqa}
    \end{minipage}
    \vspace{-3ex}
\end{figure*}

\subsection{Multi-Prototype Learning with Multiple Predicates}
\label{subsection:training}

With relation instances augmented with multiple predicate labels, we apply the multi-prototype learning strategy to train~\proposed. For each relation instance, we minimize the distance between its relation embedding $\boldsymbol{r}$ and its corresponding prototypes of the multi-labeled predicates (i.e., $\boldsymbol{\bar{c}}_{\text{gt}}$ and $\boldsymbol{\bar{c}}_{\text{aug}}$) as follows:
{\small
\begin{align}
\label{eqn:multi_proto}
\mathcal{L}_{\text{multi-proto}} = - & \lambda_i  \log \frac{\exp(\langle \boldsymbol{\bar{r}}, \boldsymbol{\bar{c}_{\text{gt}}} \rangle) / \gamma}{ \sum_{j=1}^{N_p} \exp(\langle \boldsymbol{\bar{r}}, \boldsymbol{\bar{c}}_j \rangle) / \gamma} \\ 
& - (1- \lambda_i)\log \frac{\exp(\langle \boldsymbol{\bar{r}}, \boldsymbol{\bar{c}_{\text{aug}}} \rangle) / \gamma}{ \sum_{j=1}^{N_p} \exp(\langle \boldsymbol{\bar{r}}, \boldsymbol{\bar{c}}_j \rangle) / \gamma} \nonumber
\end{align}
}
\noindent where $\boldsymbol{\bar{r}}=\textsf{Proj}(\boldsymbol{{r}})$ is the relation embedding after the projection layer is applied, $\boldsymbol{\bar{c}_{\text{gt}}}$ denotes the corresponding class prototype of the ground truth predicates, $\boldsymbol{\bar{c}_{\text{aug}}}$ denotes the prototype of augmented predicates. It is important to note that while the multi-prototype learning loss calculation considers up to two predicate classes per relation instance, our predicate augmentation approach based on sampling allows for incorporating more than two predicates by altering the augmented predicate at each batch iteration. Multi-prototype learning ensures that the embedding of a relation instance not only incorporates the original semantic of its ground-truth predicate but also embraces the semantic of the discovered latent fine-grained predicates. 

To train~\proposed, the final loss $\mathcal{L}_{\text{final}}$ is expressed as:
{\small
\begin{equation}
    \mathcal{L}_{\text{final}} = \mathcal{L}_{\text{multi-proto}} + \mathcal{L}_{\text{reg1}} + \mathcal{L}_{\text{reg2}}
\label{eqn:final_loss}
\end{equation}
}
\noindent where $\mathcal{L}_{\text{reg1}} = \sqrt{\sum_{i\neq j} (\boldsymbol{\bar{c}}_{i}^{\top} \boldsymbol{\bar{c}}_j)^2 }$ is the prototype similarity regularization~\cite{Zheng_2023_CVPR_PENET}, which aims to minimize cosine similarities between all pairs of individual prototypes, and $\mathcal{L}_{\text{reg2}}= \max \{\gamma' - \frac{1}{N_p}\sum_{i\neq j} \|\boldsymbol{\bar{c}}_{i}-\boldsymbol{\bar{c}}_{j} \|_2^2 , 0 \}$ is the prototype distance regularization~\cite{Zheng_2023_CVPR_PENET}, which increases the euclidean distance between prototypes, respectively. These regularizations serve to prevent the collapse of prototypes with semantically similar predicates. 

Note that the retrieval process and multi-prototype learning strategy are exclusively applied during the training phase. During the inference phase, we calculate the cosine similarity between each relation embedding and the set of prototypes. The predicate corresponding to the highest cosine similarity is selected as the model's prediction. 
    

\section{Experiments}
\label{sec:experiments}

\subsection{Experimental Settings}

\smallskip
\noindent \textbf{Datasets and Implementation Details.} \@
Following the prior approaches~\cite{dong2022stacked,Li_2023_ICCV_cfa,Li2021bgnn,zellers2018neuralmotif,zhang2022ietrans}, we used the benchmark datasets, VG~\cite{krishna2017visualgenome} and GQA~\cite{Hudson_2019_CVPR_gqa}. VG split contains the most frequent 150 object classes and 50 predicate classes. GQA is another vision and language benchmark, which includes top-200 object classes and 100 predicate classes. For all baselines, we adopt ResNeXt-101-FPN~\cite{resnet_101} and Faster R-CNN~\cite{faster_rcnn} as the object detector.
For~\proposed, we set the threshold $\tau=0.3$ at the reliable multi-labeled instance selection, and choose the number of retrievals $K$ by performing the grid search in $\{1,3,5,10,20 \}$. 

\smallskip
\noindent \textbf{Evaluation Protocol.} \@ We evaluate state-of-the-art models and~\proposed~on three conventional SGG tasks: (1) \textbf{Predicate Classification (PredCls)}, (2) \textbf{Scene Graph Classification (SGCls)}, and \textbf{Scene Graph Detection (SGDet)}. \textbf{PredCls} is the task that predicts the predicate classes given all ground-truth object bounding boxes and the object classes. \textbf{SGCls} aims at predicting the predicate classes given the ground-truth object bounding boxes. \textbf{SGGen} detects all entities and their pairwise predicates given an image.

\smallskip
\noindent \textbf{Metrics.} \@ We evaluate SGG models on the three metrics: (1) Recall@K (R@K) calculates the proportion of top-K predicted triplets that are in ground truth. (2) Mean Recall@K (mR@K) calculates the average recall for each predicate class, which is designed to measure the performance of SGG models under the long-tailed predicate distribution. (3) Recent SGG studies suggest that there is a trade-off between R@K and mR@K~\cite{kim2024adaptive}. Hence, recent works have focused on achieving the great F@K, where F@K calculates the harmonic average of R@K and mR@K.

\begin{table*}[t]
    \begin{minipage}{1.2\columnwidth}
    \centering
        \resizebox{1.05\textwidth}{!}{
        \centering
        \begin{tabular}{l|ccc|ccc}
        \toprule
        \multicolumn{1}{c|}{\multirow{2}{*}{\textbf{Model}}} & \multicolumn{3}{c}{\textbf{Predicate Classification}} & \multicolumn{3}{c}{\textbf{Scene Graph Classification}} \\ \cmidrule{2-7} 
        \multicolumn{1}{c|}{}                                & \textbf{R@50/100} & \textbf{mR@50/100} & \textbf{F@50/100} & \textbf{R@50/100} & \textbf{mR@50/100} & \textbf{F@50/100} \\ \midrule
        Vanilla PE-Net &    \cellcolor[HTML]{EFEFEF}64.9/67.2 & 31.5/33.8 &  42.4/45.0 & \cellcolor[HTML]{EFEFEF}39.4/40.7 & 17.8/18.9 & 24.5/25.8 \\
        \proposed~w/o select.        &  \cellcolor[HTML]{EFEFEF}64.4/66.4          & 33.4/36.4             & 44.0/47.0 & \cellcolor[HTML]{EFEFEF}38.5/39.4  & 19.6/20.9 &  26.0/27.3 \\
        \proposed~w/o IPSS &  \cellcolor[HTML]{EFEFEF}64.6/66.7          &  32.9/35.1          &  43.6/46.0 & \cellcolor[HTML]{EFEFEF}38.6/39.5 & 18.6/19.8 &    25.1/26.3      \\
         
        \proposed  & \cellcolor[HTML]{EFEFEF}62.2/64.1 & \textbf{36.2/39.1} & \textbf{45.7}/\textbf{48.6} & \cellcolor[HTML]{EFEFEF}38.2/39.1 & \textbf{20.9/22.5} & \textbf{27.0}/\textbf{28.6} \\ 
        \midrule 
        \bottomrule
    \end{tabular}
    }
    \captionof{table}{Ablation study of~\proposed.}
    \label{tab:ablation}
    \end{minipage}
    \hfill
    \begin{minipage}{.84\columnwidth}
    \centering
    \includegraphics[width=0.8\columnwidth]{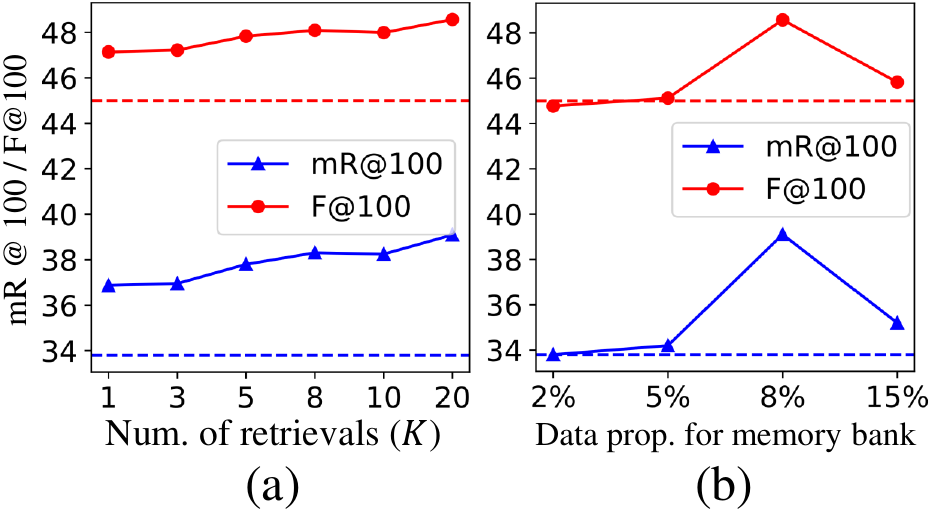}
    \vspace{-2.5ex}
    \captionof{figure}{Sensitivity analysis on hyperparameters. The dotted line represents the result of PE-Net.}
    \label{fig:sens_K}
    \end{minipage}
    \vspace{-2ex}
\end{table*}

         

\subsection{Comparisons with State-of-the-art Methods}

\smallskip
\noindent \textbf{Results on Visual Genome.} \@ We apply~\proposed~to the PE-Net backbone~\cite{Zheng_2023_CVPR_PENET}, which is specialized in capturing the relation representation space. We compare it with state-of-the-art baselines in Table~\ref{tab:main_vg}. Based on the experimental result, we have summarized the following conclusions:
\textbf{1)~\proposed~achieves state-of-the-art performance under conventional SGG tasks, particularly in terms of mR@K and F@K.} Our method not only shows better performance than the cutting-edge SGG models, such as HetSGG~\cite{yoon2023hetsgg} and SQUAT~\cite{Jung_2023_CVPRsquat} by up to 6.2\% in terms of F@K, but also achieves greater R@K. Compared to the backbone model, PE-Net, we greatly improve the performance in terms of mR@K, while slightly decreasing R@K. These results imply that~\proposed's strategy that discovers the latent fine-grained predicates through the retrieval is effective in enhancing the model's generalization capability across all predicate categories without compromising the recognition of general predicates. 
\textbf{2) Discovering the latent multi-label is a more effective strategy than data enhancement/augmentation approaches in addressing the challenges of SGG.}~\proposed~outperforms IE-Trans and CFA with the PE-Net backbone in terms of all metrics. We claim that it is attributed to the fact that while IE-Trans and CFA overlook the comprehensive understanding of general predicates to obtain the fine-grained predicates, our multi-label discovery effectively addresses both general and fine-grained predicates. This result indicates that the formulation of multi-label classification with partial annotation fundamentally addresses both the long-tailed problem and semantic ambiguity.

\noindent \textbf{Per Predicate Comparison.}\@ In Figure~\ref{fig:perclass_vg}, we conduct a detailed comparison between~\proposed~utilizing the PE-Net backbone, and the vanilla PE-Net by investigating the per predicate performance analysis. Our analysis reveals that~\proposed~significantly enhances performance on fine-grained predicates such as ``lying on,'' ``walking in,'' ``made of,'' and ``says,'' which are entirely overlooked by the vanilla PE-Net. This improvement indicates that~\proposed~excels in capturing the nuanced visual cue to predict fine-grained predicates, demonstrating the effectiveness of discovering the latent predicates.

\noindent \textbf{Results on GQA.}\@ In Table~\ref{tab:main_gqa}, we observed that~\proposed~greatly improves mR@K and F@K compared to the strong baseline, PE-Net, showcasing the effectiveness of~\proposed~on a large dataset, which includes numerous predicate categories. Similarly, Figure~\ref{fig:perclass_gqa}~demonstrates~\proposed's ability to improve the recognition of fine-grained predicates, which PE-Net tends to overlook. These results imply that~\proposed~effectively discovers the latent fine-grained predicates even when using the dataset with numerous predicate categories.


\subsection{Ablation Studies}

\smallskip
\noindent \textbf{Module Component Analysis.}\@ We conduct the ablation study of~\proposed~to evaluate the impact of each component. We ablate two components in~\proposed. For~\proposed~\textsf{w/o select.}, we skip the reliable multi-labeled instance selection, which determines whether a relation instance is single-labeled or multi-labeled. For~\proposed~\textsf{w/o IPSS}, we replace the inverse propensity score with a constant in the unbiased augmentation module. In Table~\ref{tab:ablation}, we obtained the following observations: 1)~\proposed~\textsf{w/o select.} still shows better performance than PE-Net in terms of mR@K and F@K. That is, the strategy, which roughly assigns pseudo-labels for all relations based on retrieved instances, can enhance the performance on fine-grained predicates. We argue that it is attributed to the superiority of the retrieval-augmented approach, which discovers semantically similar predicates using the memory bank. 2)~\proposed~\textsf{w/o IPSS} decreases mR@K and F@K compared to~\proposed. This implies that inverse propensity score-based sampling is important for boosting the performance on fine-grained predicates. 3) Combining all components allows for the best generalization capability of~\proposed.

\smallskip

\noindent\textbf{Analysis on the Number of Retrievals.} We analyze the effect of $K \in \{1, 3, 5, 8, 10, 20 \}$ under PredCls task. In Figure~\ref{fig:sens_K}.(a), we observe that~\proposed~consistently outperforms PE-Net, indicating that utilizing even one near neighbor is helpful for enhancing the performance of PE-Net.~\proposed~achieves the best performance at $K=20$, implying that referring more neighbors is helpful for enhancing the performance on fine-grained predicates.

\smallskip

\noindent\textbf{Analysis on the Size of Memory Bank.} We anavlyze the impact of the memory bank size and vary the number of relation instances stored per unique triplet, testing with $\{1, 5, 10, 100\}$ instances for each unique triplet. These quantities represent approximately $\{2\%, 5\%, 8\%, 15\%\}$ of the entire training instances, respectively. In Figure~\ref{fig:sens_K}.(b), we observed that~\proposed~generally outperforms the vanilla PE-Net, and attains the best performance when the memory bank comprises $8\%$ of the training instances. This result suggests that leveraging a small portion of the training dataset can significantly improve the performance of SGG models. Moreover, the compact size of the memory bank not only enhances its efficiency but also contributes to the scalability of~\proposed.

\smallskip

\noindent \textbf{Performance of Retrieval Strategy.} As there are no ground truth retrieval labels, it is difficult to measure the accuracy of retrievals. Hence, we conducted a human evaluation of the retrieval process on 100 sampled instances, and obtained $84.20\%$ accuracy. This implies that our approach performs accurate retrieval and effectively augments the multi-labels. 

\smallskip

\noindent \textbf{Complexity Analysis.} Note that the inference time for~\proposed~is equivalent to that of PE-Net. At the training time, we observed that~\proposed~ requires 1.3 times training time of PE-Net, which implies that the computational cost of the retrieval process is manageable.

\section{Conclusion}
Although research on fine-grained SGG has garnered its attention, fundamental challenges still remain in the prior framework. They either suffer from capturing the fine-grained predicates due to the SGG nature, which encompasses the long-tailed problem and semantic ambiguity, or overlook the general predicates by re-annotating the general predicates as fine-grained predicates. In this work, we first reformulate the SGG task as multi-label classification problem with partial annotation by discovering potential fine-grained predicates in the training data. To this end, we propose a novel retrieval-augmented SGG framework (\proposed), which enhances original labels to multi-labels through a retrieval process.~\proposed~effectively discovers latent fine-grained predicates in the training dataset, resulting in substantial improvements across overall predicates. Our works presents a new perspective in SGG research by considering the nature of natural language, where multiple predicates convey the same single relationship.

\section*{Acknowledgement}

This work was supported by National Research Foundation of Korea(NRF) funded by Ministry of Science and ICT (NRF-2022M3J6A1063021, 10\%, No. RS-2024-00341514, 10\%), Institute of Information \& Communications Technology Planning \& Evaluation(IITP) grant funded by the Korea government(MSIT) (RS-2023-00216011, Development of artificial complex intelligence for conceptually understanding and inferring like human, 79\%) and (No. 2019-0-00079, Artificial Intelligence Graduate School Program, Korea University, 1\%).

\bibliography{aaai25}

\newpage
\appendix

\clearpage

\section{Implementation Details}

For a fair comparison, we employ the pre-trained ResNeXt-101-FPN backbone and Faster R-CNN detector~\cite{faster_rcnn} by following the setting of previous studies~\cite{zellers2018neuralmotif,zhang2022ietrans,Li_2023_ICCV_cfa}. For~\proposed, we search the number of retrievals $K$ by conducting grid search in $\{1,3,5,10,20 \}$ using holdout set, and select the model with the best F@K performance. We set threshold $\tau=0.3$. Moreover, when we assign multi-labels for the background relationships, we skip the reliable multi-labeled instance selection phase and only perform unbiased multi-label augmentation based on inverse propensity score-based sampling. We employ the NVIDIA A6000 48GB device for our experiments.

\section{Additional Analysis on RA-SGG}
\label{appendix:section_analysis}

\subsection{Complexity Analysis}
\label{appendix:sec_complexity}

\begin{table}[h]
\centering
\scalebox{.82}{
\begin{tabular}{l|c|}
\hline
Method & Time (s) per 100 batch iterations \\
\hline
$\text{PENet}^{\dagger}$ \cite{Zheng_2023_CVPR_PENET}{\tiny{CVPR'23}} & 20.42 \\
$\textbf{\proposed}$ & 26.03 \\
 \bottomrule
\end{tabular}
}
\caption{Time (s) Per 100 Batch Iterations.~\proposed~employs the PE-Net backbone.}
\label{tab:time_complexity}
\end{table}

We conduct the complexity analysis to verify that using a memory bank in~\proposed~requires affordable computational cost. In this experiment, we set $K=5$ and use the memory bank with $8\%$ of the training dataset (i.e., about $90,000$ relation instances) in the retrieval process. We employ the NVIDIA A6000 device for the experiment. Table~\ref{tab:time_complexity} shows the required times for training PE-Net and~\proposed, respectively. We observed that the training time of~\proposed~is about 1.3 times that of PE-Net, implying that the computational cost of the retrieval process is manageable. During the testing phase, the inference time for~\proposed~is equivalent to that of PE-Net.

\begin{figure}[t!]
    \centering
    \includegraphics[width=1.0\columnwidth]{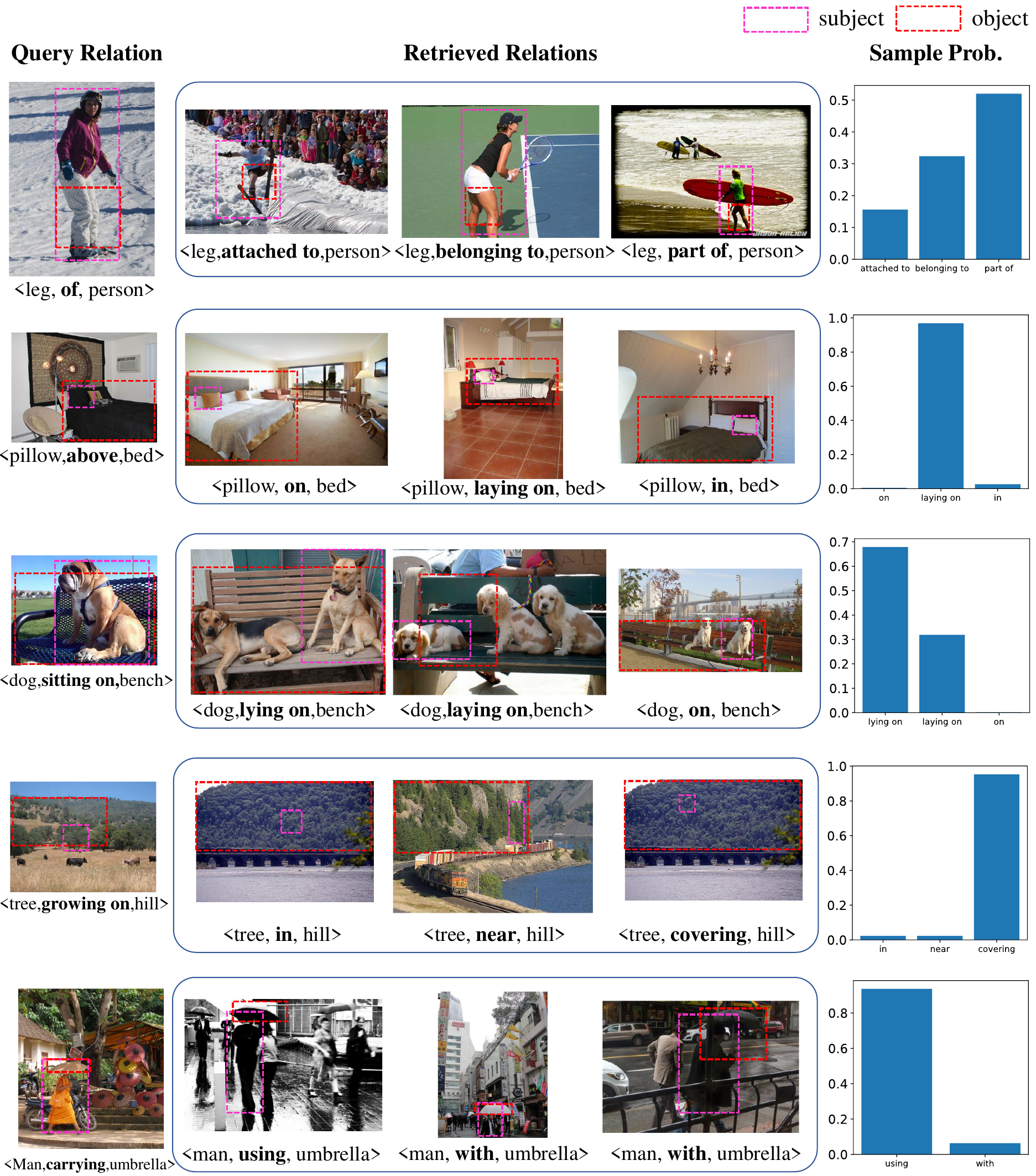}
    \caption{Examples of query relation, retrieved relations, and the sample probabilities in~\proposed. Given query relation, we retrieve top-$3$ similar instances from the memory bank. Then, we compute the sample probability based on the inverse propensity score of retrieved instances. We generate multi-labels by sampling predicate labels based on the sample probability.}
    \label{fig:qualitative}
\end{figure}

\subsection{Qualitative Analysis}

\smallskip
\noindent \textbf{Examples in Retrieval Process.} In Figure~\ref{fig:qualitative}, we randomly select  examples of the query relations and their retrieved instances, and visualize them. Given the query relations,~\proposed~finds the semantically similar relation instances from the memory bank. We can observe that the predicate classes of retrieved instances can be potential relationships of the query relations. Moreover, the sample probability computed by the inverse propensity scores of retrieved instances encourages the generation of pseudo-labels on fine-grained predicate classes.

\smallskip

\begin{figure}[t!]
  \centering
  \includegraphics[width=0.9\linewidth]{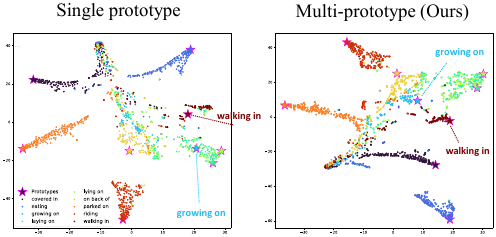}
   \caption{T-sne of single and multi-prototype learning.}
   \label{fig:tsne}
\end{figure}

\noindent \textbf{Visualization of Visual Semantic space.} In Figure~\ref{fig:tsne}, single-prototype learning shows prototypes for tail classes like ``growing on'' and ``walking in'' distantly positioned from their samples, indicating ineffective learning on tail classes. Conversely, multi-prototype learning trains prototypes closer to the samples for these classes, suggesting enhanced prediction of fine-grained predicates.

\subsection{Analysis on Various Retrieval-Augmented Methods} 

We analyze various strategies for utilizing retrieved instances to obtain fine-grained SGG. Specifically, when provided with a query relation $(\boldsymbol{r}^{\text{query}}, \mathbf{p}^{\text{query}})$ and retrieved instances $\{ (\boldsymbol{r}^{\text{ret}}_1, \mathbf{p}^{\text{ret}}_1), ..., (\boldsymbol{r}^{\text{ret}}_K, \mathbf{p}^{\text{ret}}_K) \}$, we suggest three ways to utilize the retrieved instances. 1) \textsf{Feat Aug}: Directly enhance the query relation embedding with retrieved predicates by using MLP without label augmentation. (i.e., $\boldsymbol{r}^{\text{query}} \leftarrow \boldsymbol{r}^{\text{query}} + \textsf{MLP}(\{\mathbf{p}^{\text{ret}}_i\}_{i=1}^K)$). 2) \textsf{Label Aug (RA-SGG)}: Augment the predicate label of query relations using our proposed method (i.e., $\mathbf{p}^{\text{query}} \leftarrow \lambda \mathbf{p}^{\text{query}} + (1-\lambda) \mathbf{p}^{\text{aug}}$). 3) \textsf{Mix-up}: Augment both the relation embedding and the predicate labels of query relations by using mix-up~\cite{zhang2018mixup} (i.e., $\boldsymbol{r}^{\text{query}} \leftarrow \lambda \boldsymbol{r}^{\text{query}} + (1-\lambda) \boldsymbol{r}^{\text{aug}}$ and $\mathbf{p}^{\text{query}} \leftarrow \lambda \mathbf{p}^{\text{query}} + (1-\lambda) \mathbf{p}^{\text{aug}}$). 

In Table~\ref{tab:retrieval_way}, we observed that \textsf{FeatAug} and \textsf{Mixup} lead to a decrease in mR@K while enhancing R@K. This outcome diverges from our objective of generating a fine-grained scene graph. The observed phenomenon can be attributed to the integration of information about tail classes into the features, which inadvertently create a shortcut to general predicates. For instance, when the model learns to predict ``on'' based on the retrieved ``walking on,'' it tends to adopt shortcut patterns (i.e., ``walking on'' $\rightarrow$ ``on''), consequently enhancing the performance of head classes. In contrast, \textsf{Label Aug} shows a great increase in both mR@K and F@K metrics, indicating that~\proposed's strategy, which exclusively augments labels, is particularly effective for enhancing the fine-grained predicate predictions.

\begin{table}[t!]
\centering
\resizebox{.41\textwidth}{!}{
\centering
\begin{tabular}{l|ccc}
\toprule
\multicolumn{1}{c|}{Task: PredCls}                                & \textbf{R@50/100} & \textbf{mR@50/100} & \textbf{F@50/100} \\ \midrule

Vanilla PE-Net & \textbf{64.9/67.2}         & 31.5/33.8     & 42.4/45.0  \\
\textsf{Feat Aug}  & \textbf{68.3/70.3}   & 20.9/22.8            & 32.0/34.5          \\
\textsf{Label Aug} (\proposed)  & 62.2/64.1         & \textbf{36.2/39.1}          & \textbf{45.8/48.6}         \\
\textsf{Mix-up} & 67.3/69.4 & 25.9/28.2 & 37.4/40.1 \\

\bottomrule
\end{tabular}
}
\captionof{table}{Results on VG when various retrieval-augmented methods are applied.}
\label{tab:retrieval_way}
\end{table}

\section{Limitation}

In our study, we demonstrate the effectiveness of retrieval-augmented SGG, which discovers the latent fine-grained predicates. Although~\proposed~is restricted to the case when the memory bank is populated with training instances, our retrieval-based approach could potentially benefit from further extension by incorporating external datasets as additional sources of information. We regard this avenue for future exploration.

\end{document}